\documentclass[10pt,twocolumn,letterpaper]{article}

\usepackage[pagenumbers]{cvpr} % To force page numbers, e.g. for an arXiv version

\usepackage{pifont,xcolor}  
\usepackage{multirow}
\usepackage[table,xcdraw]{xcolor}
\usepackage{csquotes}
\usepackage[normalem]{ulem}
\usepackage{arydshln}

\usepackage{booktabs}
\usepackage{siunitx}
\definecolor{cvprblue}{rgb}{0.21,0.49,0.74}
\usepackage[pagebackref,breaklinks,colorlinks,allcolors=cvprblue]{hyperref}

\def\confName{CVPR}
\def\confYear{2026}

\title{Probing and Bridging Geometry–Interaction Cues for Affordance Reasoning in Vision Foundation Models}

\author{
Qing Zhang$^{1}$ \quad
Xuesong Li$^{1,2}$\textsuperscript{\footnotemark[2]} \quad
Jing Zhang$^{1}$\textsuperscript{\footnotemark[2]} \\[0.5em]
$^{1}$The Australian National University \quad
$^{2}$CSIRO\\[0.5em]
{\tt\small
u7561400@anu.edu.au \quad
xuesong.li@anu.edu.au \quad
jing.zhang@anu.edu.au
}
}

\begin{document}
 \maketitle
 
\renewcommand{\thefootnote}{\fnsymbol{footnote}}
\makeatother
\footnotetext[2]{Corresponding authors.\hspace{80pt}}

 \begin{abstract}
% affordance capability emergence of the current large foundation model --> how to investigate it --> how to use the insight

What does it mean for a visual system to truly understand affordance? We argue that this understanding hinges on two complementary capacities: geometric perception, which identifies the structural parts of objects that enable interaction, and interaction perception, which models how an agent's actions engage with those parts. To test this hypothesis, we conduct a systematic probing of Visual Foundation Models (VFMs). We find that models like DINO inherently encode part-level geometric structures, while generative models like Flux contain rich, verb-conditioned spatial attention maps that serve as implicit interaction priors. Crucially, we demonstrate that these two dimensions are not merely correlated but are composable elements of affordance. By simply fusing DINO's geometric prototypes with Flux's interaction maps in a training-free and zero-shot manner, we achieve affordance estimation competitive with weakly-supervised methods. This final fusion experiment confirms that geometric and interaction perception are the fundamental building blocks of affordance understanding in VFMs, providing a mechanistic account of how perception grounds action. Our code will be available at: \href{https://github.com/QZhang2111/Probing_Bridging_Affordance}{Probing and Bridging Affordance}
\end{abstract}

\begin{figure}[!t]
  \centering
  \includegraphics[width=\linewidth]{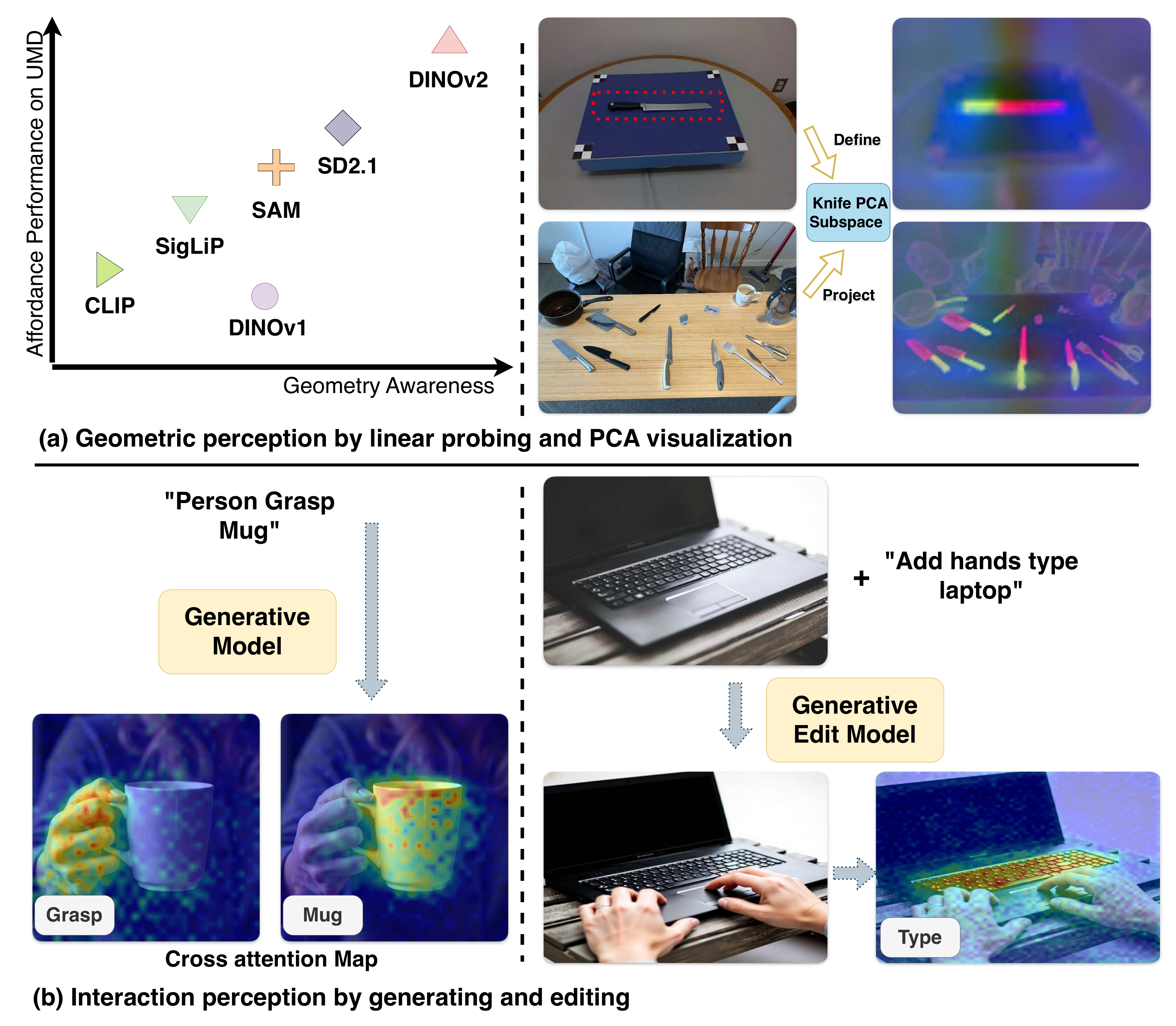   }
        \caption{ \textbf{ From geometry to interaction: uncovering affordance perception in visual foundation models.} Affordance understanding emerges from the fusion of geometric structure and generative interaction priors.
        (a) Models with stronger geometric awareness yield richer part-level representations.
        (b) Generative models reveal verb-conditioned attention that naturally localizes interaction regions without supervision. 
        % \JZ{this fig is not discussed in the intro section}
            }
    
  \label{fig:sec1/overview}
\end{figure}    
 \section{Introduction}
\label{sec:intro}

Visual affordance describes how objects in a scene can be acted upon, serving as a bridge between visual perception and embodied action~\cite{gibson2014theory,do2018affordancenet,sawatzky2017weakly,nguyen2017object}. By encoding affordance within visual representations, models can infer not only what objects are, but also how they can be used. Recent studies have exploited this concept in robotic manipulation, spatial reasoning, and vision-language navigation~\cite{zhang2024affordance,li2025learning,yuan2024robopoint,bahl2023affordances}, showing its importance in linking perception to action.
Yet despite these advances, little is known about what internal capability enables such understanding. In other words, \textit{what does it mean for a visual system to truly “understand” affordance?}

Research on visual affordance has progressed along three main paradigms, each grounded in distinct sources of evidence. \textbf{Fully-supervised} methods learn geometric patterns from dense pixel-level annotations, achieving precise region predictions at the cost of calability~\cite{do2018affordancenet,nguyen2017object,thermos2020deep,myers2015affordance}.
\textbf{Weakly supervised} approaches infer affordance from indirect behavioral cues such as human-object interactions, yet their spatial precision remains coarse~\cite{fang2018demo2vec,nagarajan2019grounded,liu2022joint,chen2023affordance,li2023locate,luo2022learning}.
\textbf{Open-vocabulary} methods leverage semantic associations from large-scale image-text pretraining, enabling generalization but often conflating function with category-level semantics~\cite{li2024one,tong2024oval,chen2025maskprompt,qian2024affordancellm,jiang2025affordancesam,Wang_2025_CVPR}.
Consequently, we still lack a unified view of the fundamental capability for affordance, independent of its supervision. Visual Foundation Models (VFMs), which internalize diverse visual knowledge, offer a new lens to address this. They allow us to directly probe for core capacities, motivating our central question: What \emph{core capabilities} enable affordance understanding, and how are they manifested within VFMs?

% Overall, these paradigms respectively emphasize \emph{geometric}, \emph{interaction}, and \emph{semantic} cues, yet lack a unified perspective for understanding what core competence constitutes affordance perception, \JZ{particularly how existing VFMs contribute such cues for affordance estimation.}  
% Although VFMs are now widely adopted as feature extractors, little is known about whether they have spontaneously developed affordance-relevant representations~\cite{tang2023emergent,el2024probing,kim2024beyond,deng2025emerging} \JZ{this is over-claim. VFMs have been explored for affordance in the open-vocabulary setting. Their limitation is that they only highlight the overall representativeness of those VFMs. Differently, you explore capacities of VFMs for affordance estimation from two main dimensions, which you believe are the main factors for affordance. Please rephrase this part}.  
% This raises a central question: what \emph{core capability} should a model truly learn from affordance tasks, and how does it manifest within existing VFMs?

To answer this, we revisit the core tenet of affordance: it is not a property of the object alone, but a possibility for interaction between an agent and its environment~\cite{gibson2014theory}.
We therefore posit that understanding affordance demands two complementary capabilities: \textbf{geometric perception}—identifying the object parts and spatial layouts that enable interaction, and \textbf{interaction perception}—modeling how an agent engages with those structures. \textit{A chair affords sitting or leaning because its seat and backrest jointly support the body, while the same chair affords grasping or lifting through its armrests or legs for the hands}, showing that affordance emerge from the coupling between the chair’s geometry and the agent’s action capabilities.
This dual perspective unifies prior approaches that either emphasize learning geometric patterns of affordance~\cite{do2018affordancenet,nguyen2017object,thermos2020deep,myers2015affordance} or human–object interaction relations~\cite{fang2018demo2vec,nagarajan2019grounded,liu2022joint,chen2023affordance,li2023locate,luo2022learning}.
At the same time, it provides a principled lens for probing how affordance-related capabilities may emerge within VFMs.

% To address the above questions, we propose a mechanistic analytical perspective:
% the capability of affordance can be operationally characterized as the combination of \textbf{geometric perception} and \textbf{interaction perception}. \JZ{you need to provide more evidence why these two dimensions matter, not just because the existing solutions have explored them. This is to answer your first question. For example, you can refer to the early affordance papers or the concept of affordane, or even give an example}
% \textbf{Geometric perception} concerns a model’s ability to organize and distinguish object parts, shapes, and spatial structures—determining whether it can perceive boundaries and configurations within a scene.
% \textbf{Interaction perception} captures the spatial responsiveness to \enquote{where and how} an agent acts upon an object, reflecting sensitivity to action-related cues.
% These two dimensions align with the main lines of affordance research:
% fully supervised tasks learning geometric distributions from dense annotations,
% and weakly supervised tasks learning interaction regions from behavioral or demonstration data(Fig~\ref{fig:sec1/paradigm}).
% Analyzing VFMs from these complementary \JZ{be specific, why they are complementary} perspectives reveals how affordance-related capabilities emerge internally and exposes their potential geometric–interaction complementarity across tasks. 

To empirically ground this framework, we investigate how the two proposed dimensions—geometry and interaction—relate to affordance understanding and how they are manifested within VFMs as visualized in Fig.~\ref{fig:sec1/overview}. 
Across diverse models, stronger geometric awareness~\cite{el2024probing} leads to higher affordance segmentation accuracy, while adding depth or surface-normal cues further strengthens this link (Sec. ~\ref{sec3: Geometry}). 
Models that inherently encode interaction priors also demonstrate comparable affordance estimation performance with weakly supervised methods, confirming that both dimensions are empirically tied to affordance capabilities (Sec. ~\ref{sec3:interaction}).
Building on these observations, we further explore how VFMs express these two capabilities. 
On the geometric front, we find that models best at affordance segmentation organize visual features into part-level structures, revealing object geometry as the foundation of actionable perception.
In parallel, we discover that generative models learn a verb-conditioned spatial attention that links actions to object regions, allowing these interaction priors to be explicitly extracted as spatial guides.
To validate their complementarity, we fuse these cues into a training-free framework. This composition enables zero-shot affordance estimation competitive with weakly-supervised methods, demonstrating that these core capacities exist as composable elements within VFMs.
The main contributions of this work are summarized as follows:
\begin{itemize}
\item We propose a dual-dimensional framework decomposing visual affordance into complementary \textbf{geometric} and \textbf{interaction} perception, offering a unified basis for analysis beyond supervision-specific paradigms;
\item We conduct the first systematic probing of affordance capabilities across diverse VFMs, revealing that geometric awareness enables part-level separability, while generative models encode verb-conditioned spatial interaction priors;
\item We present, to the best of our knowledge, the first work that exploits internal cross-attention in generative models as a rich source of \textbf{training-free spatial priors} for affordance estimation.
\item We demonstrate their composability via a training-free fusion, achieving zero-shot estimation competitive with weakly-supervised methods.
\end{itemize}

\begin{table}[t]
    \centering
    \resizebox{\linewidth}{!}{
    \begin{tabular}{llll}
        \toprule
        \textbf{Model} & \textbf{Architecture} & \textbf{Supervision}  \\
        \midrule
        DINO~\cite{caron2021emerging}    & ViT-B/16   & SSL     \\
        DINOv2~\cite{oquab2023dinov2}    & ViT-B/14   & SSL       \\
        DINOv3~\cite{simeoni2025dinov3}  & ViT-B/16   & SSL      \\
        CLIP~\cite{ilharco_gabriel_2021_5143773}      & ViT-B/16  & VLM       \\
        SigLIP~\cite{zhai2023sigmoid}           & ViT-B/16   & VLM               \\
        SAM~\cite{kirillov2023segment}   & ViT-B/16   & Segmentation         \\
        StableDiffusion~\cite{rombach2022high}   & UNet       & Generation        \\
        Flux~\cite{labs2025flux}        & DiT       & Generation     \\
        
        \bottomrule
    \end{tabular}
    }
    \caption{\textbf{Probed Visual Models.} We used supervised classification from Probing3D~\cite{el2024probing} to explore and analyze these models.}
    \label{tab:model_comparison}
\end{table}
\section{Related work}
\label{sec:related_work}

\begin{figure}[!t]
  \centering
\includegraphics[width=\linewidth]{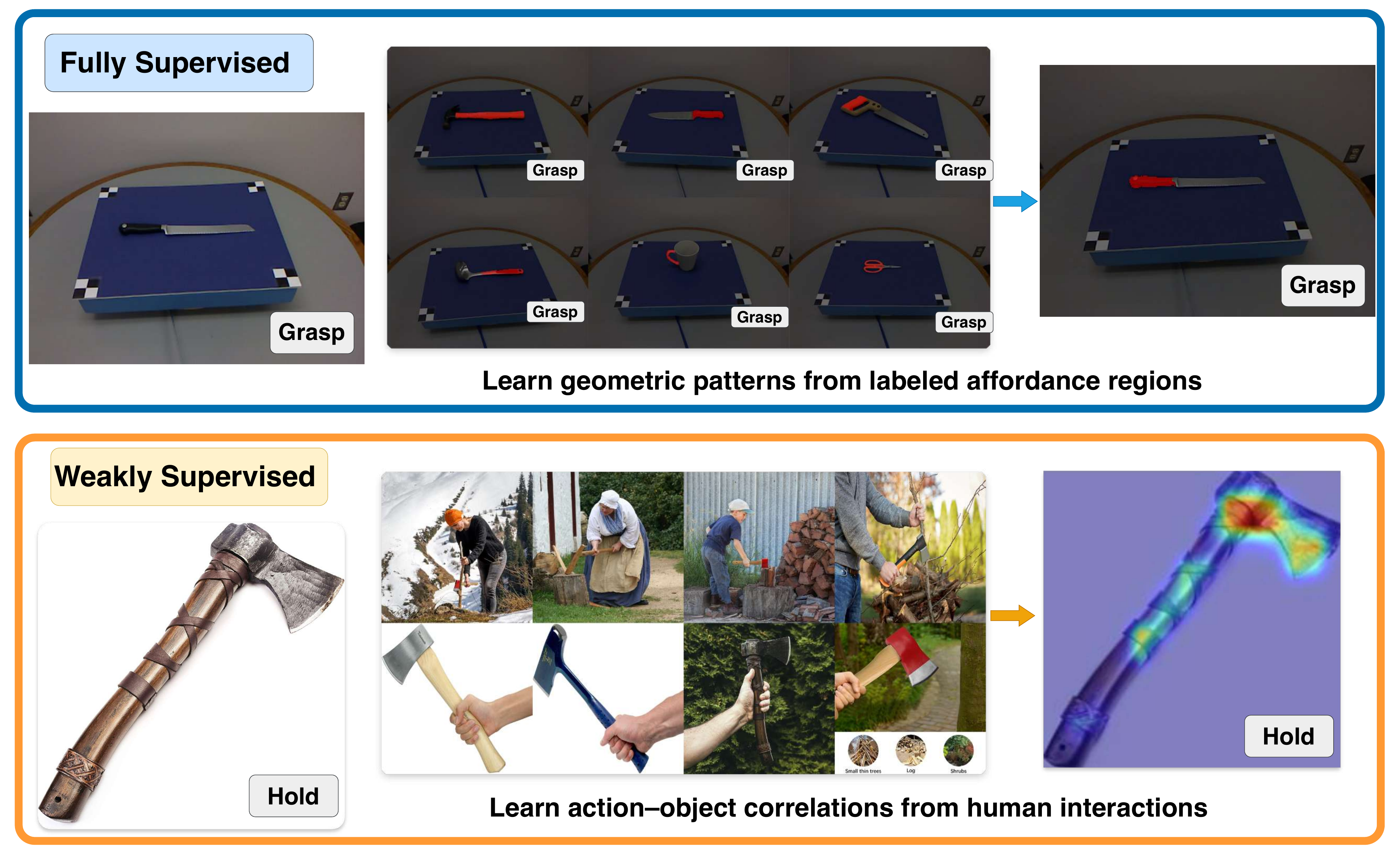}
    \caption{\textbf{Learning Affordances: From Geometry to Interaction.} Fully-supervised methods learn geometric structures from pixel masks; weakly-supervised ones infer interactions from human-object imagery.}
  \label{fig:sec1/paradigm}
\end{figure}

\noindent\textbf{Visual Affordance Learning.} 
Existing work seeks to localize object parts that enable actions, and generally follows three supervision paradigms, each emphasizing different evidence.
\textit{Fully supervised methods} learn geometric patterns from dense affordance annotations achieving reliable performance under fixed category–geometry distributions\cite{do2018affordancenet,Deng_2021_CVPR,xu2022partafford,myers2015affordance,thermos2020deep}(Fig.~\ref{fig:sec1/paradigm} $1^{st}$ row).
However, the learned priors remain tightly coupled to labeled categories and parts, limiting generalization across objects and semantic domains.
\textit{Weakly supervised methods} infer affordance regions from indirect behavioral cues—such as image-level action labels, HOI videos, or demonstration alignment—by transferring interaction evidence from human actions to static objects \cite{fang2018demo2vec,nagarajan2019grounded,liu2022joint,luo2022learning,li2023locate,chen2023affordance} (Fig.~\ref{fig:sec1/paradigm} $2^{nd}$ row).
While they effectively capture action-object correlations, their spatial localization remains coarse, though recent works improve consistency using frozen ViT features and prototype-based regularization \cite{cui2023strap,xu2024weakly,xu2025weaklysupervised,Wang_2025_CVPR}. \textit{Open-vocabulary methods} leverage vision–language alignment or multimodal generation to map arbitrary verbs or actions to image regions, achieving one-shot/zero-shot affordance prediction \cite{li2024one,tong2024oval,chen2025maskprompt,qian2024affordancellm,jiang2025affordancesam,lee2025affogato}.
While scalable to unseen objects, their predictions often rely on semantic associations rather than grounded interaction cues, rendering performance highly dependent on prompt and retrieval quality\cite{van2024open,chen2024worldaffordaffordancegroundingbased}. Overall, these paradigms respectively emphasize geometric, interaction, and semantic evidence. However, they lack a unified framework and rarely examine whether such affordance-relevant capabilities already emerge within VFMs—a gap we explore next.

% \JZ{how about draw a figure to explain clearly the main pipeline of those existing solutions, and yours. thus highlight the innovation}

\noindent\textbf{Visual Foundation Models for Affordance Tasks.}
% \JZ{typically, visual foundation models can be roughly divided into discriminative models (e.g. self-supervised discriminative models, DINO, CLIP, MAE, SimCLR) and generative models (diffusion models (stablediffusion), autogressive models (GPT-4v)). Maybe re-organize this part according to the above two different category}
With the emergence of large-scale pretrained models, visual affordance research increasingly builds upon VFMs to obtain transferable and generalizable visual representations\cite{qian2024affordancellm,jiang2025affordancesam,lee2025affogato}.
These models can be broadly divided into discriminative and generative types, corresponding to representation learning and image synthesis paradigms in visual modeling. 
\textit{Discriminative Models} learn transferable visual representations under various forms of supervision.
Self-supervised models trained through contrastive or reconstruction objectives develop strong spatial organization and 3D awareness~\cite{amir2021deep,el2024probing}.
They have been widely adopted as backbones for affordance estimation~\cite{li2024one,cui2023strap,xu2025weaklysupervised}, yet their affordance-relevant capabilities remain underexplored and not well characterized.
Vision–language models leverage large-scale image–text alignment to enable one-shot and zero-shot affordance prediction~\cite{zhou2023zegclip,deng2025emerging}.
While their open-vocabulary recognition supports scalability across novel actions and objects, their predictions often rely on high-level semantics, with limited fine-grained correspondence between verbs and object parts~\cite{van2024open,chen2024worldaffordaffordancegroundingbased}.
Segmentation-supervised models trained on large-scale mask annotations provide reliable boundary and instance priors~\cite{jiang2025affordancesam}.
They are primarily used to generate pseudo-labels or auxiliary supervision in affordance pipelines, rather than to perform affordance estimation directly. \textit{Generative Models} learn to synthesize visual content from large-scale text–image data, modeling the joint distribution of objects, scenes, and relations through reconstruction or diffusion objectives~\cite{rombach2022high}.
They have been primarily applied to text-to-image generation, editing, and controllable synthesis, where the focus lies on visual fidelity and compositional coherence.
To date, such models have rarely been examined in the context of affordance learning~\cite{tang2023emergent,kim2024beyond}.
To the best of our knowledge, this work provides the first systematic investigation into affordance-relevant interaction priors within generative models, offering a new perspective on how action-related cues may emerge through the generative process.

% \textbf{DINO} and \textbf{DINOv2}, known for their strong separability at the part and shape levels \cite{amir2021deep,el2024probing}, are widely used in weakly supervised and open-vocabulary settings.  
% By freezing features or selecting prototypes \cite{cui2023strap,xu2025weaklysupervised}, these models enhance geometric perception and structural coherence. \JZ{you mentioned that those models have already explored geometric perception of affordance, then you need to answer what makes your exploration different}  
% \textbf{CLIP} and \textbf{SigLIP} provide open-vocabulary semantic guidance through image–text alignment \cite{zhou2023zegclip,deng2025emerging}, supporting retrieval, prompt-based inference, and region grounding.  
% However, their fine-grained correspondence between verbs and object parts remains uncertain \cite{van2024open,chen2024worldaffordaffordancegroundingbased}.  
% \textbf{SAM}, as a universal segmentation model, offers instance masks or pseudo-labels to weakly supervised and open-vocabulary methods \cite{jiang2025affordancesam}, significantly reducing the difficulty of pixel-level learning.  

% overview why we investigation from this two dimension.
\begin{figure*}[!t]
  \centering
  \includegraphics[width=\linewidth]{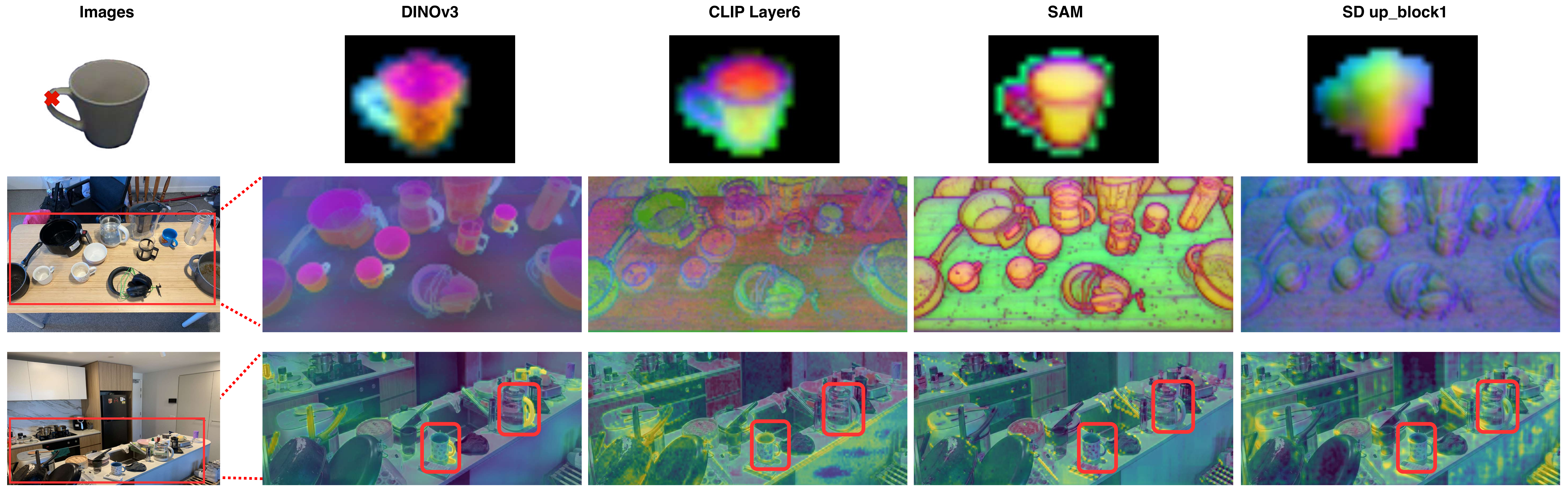}
      \caption{ \textbf{Distinct geometric representations across Visual Foundation Models. }
        We project complex scenes into the PCA subspace of a reference object (a mug) to visualize and compare geometric representations.
        DINOv3 produces part-level structures that generalize across scenes and materials, as confirmed by cosine similarity responses from a mug-handle patch, while CLIP, SAM, and Stable Diffusion emphasize semantics, edges, or smooth surface continuity instead.
        }
  \label{fig:sec2/cos_sim}
\end{figure*}

\begin{figure}[!t]
  \centering
  \includegraphics[width=\linewidth]{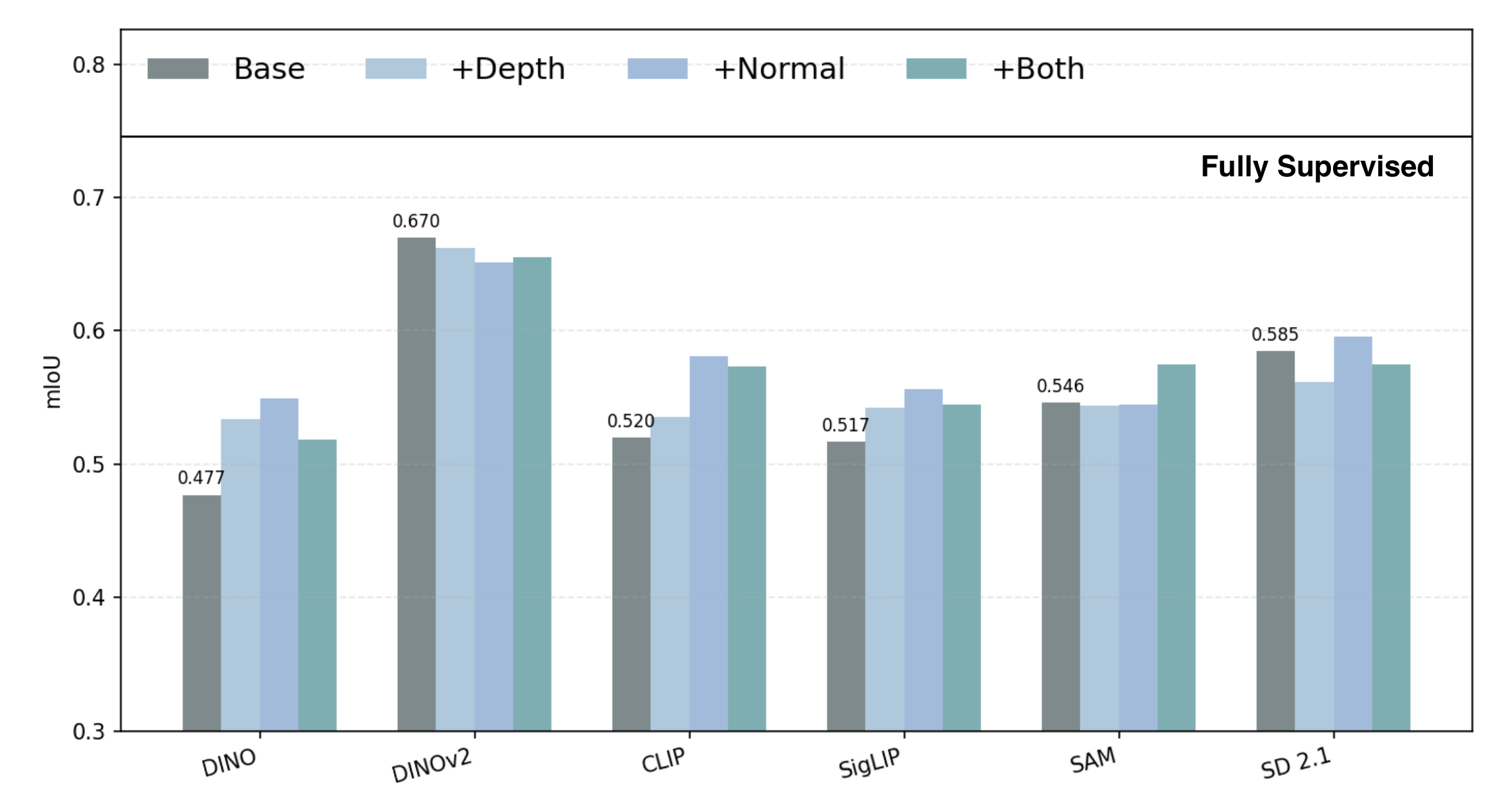}
  \caption{\textbf{ Geometric awareness predicts affordance perception ability.} We evaluate visual models under different supervision paradigms using linear probing on the UMD dataset~\cite{myers2015affordance,do2018affordancenet}. Models with stronger geometric awareness~\cite{el2024probing}—achieve higher mIoU in affordance segmentation, while others improve notably when augmented with depth or normal cues.
  % \XS{Plot the bar chart with deep color}
  }
  \label{fig:sec2/linear_prob}
\end{figure}

%\section{Mechanistic Investigation of Visual Affordance }
% \section{Systematic Investigation of Visual Affordance}
\section{Visual Affordance Investigation}
\label{sec:two_dimentsion}

% Previous solutions either explore affordance in a fully supervised way or a weakly supervised way. Vision-language-models (VLMs) make it possible to achieve open-vocabulary affordance estimation \JZ{this is redundant}. However, we argue that all those solutions focus more on geometry relationship understanding, failing to discovering the main concept for affordance, namely interaction~\XS{This is a strong position, VLM doesn't use the geometry relation and rely more on interaction. Here, we just need to provide a high-level background for geometric and interaction in a natural tone.} \JZ{agreed}. From our perspective, affordance indicates the interaction of two different instances. In this case, it's also necessary to investigate how interaction, especially the capabilities of interaction modeling of existing vision foundation models and their potential for affordance estimation. In this case, we investigate affordance from two main dimensions, namely geometry level and interaction understanding level. 
Building on our dual-dimensional framework, we now investigate how the two fundamental capabilities—geometry and interaction—relate to affordance understanding and how they are manifested within VFMs.
Rather than introducing new task-specific architectures, our goal is to empirically ground the proposed dimensions by probing existing VFMs across geometric and interactive reasoning. 

Sec.~\ref{sec3: Geometry} examines the geometric dimension, testing whether geometric awareness correlates with affordance estimation and analyzing how different VFMs encode geometric structures that support action possibilities.
Sec.~\ref{sec3:interaction} focuses on the interaction dimension, exploring whether generative models inherently encode verb-conditioned interaction priors that can guide affordance estimation without explicit supervision.

\subsection{Geometry}
\label{sec3: Geometry}
Our investigation confirms that geometric representation forms the structural substrate of affordance learning. We first establish a direct quantitative link between a model's geometric awareness and its affordance segmentation performance. We then delve into the internal representations of leading models, comparing how different VFM families encode geometry and revealing that part-level organization is a hallmark of strong geometric perception for action.

\vspace{3pt}
\noindent\textbf{Observation 1.1. Affordance Correlates with Geometric Representation.}
% We begin by testing whether a model’s geometric competence directly influences its affordance prediction ability.
% Following the Probing3D protocol \cite{el2024probing}, we evaluate several VFMs (Table~\ref{tab:model_comparison})\XS{move table close to this page} using pixel-level linear probing. \XS{Explain more about its dataset, supervision, linear layers, loss function, and training steps, etc. in the technical section.}
% Models with stronger geometric sensitivity in Probing3D rankings achieve higher mean IoU on affordance segmentation, confirming a strong correlation between geometric representation and affordance separability.
% When explicitly adding depth and surface-normal cues estimated by Metric3Dv2 \cite{hu2024metric3d}, performance consistently improves across all models, particularly for CLIP and SigLIP, whose representations are otherwise dominated by semantic alignment.
% This pattern confirms that geometry is not an incidental byproduct but a prerequisite for affordance learning. \XS{explain why these depth and surface-normal cues are not helpful for DINOv2?}
To quantitatively validate that geometric awareness underpins affordance understanding, we adopt a controlled linear probing framework following the Probing3D protocol~\cite{el2024probing}.
The evaluation is conducted on the UMD dataset, which contains seven affordance categories with 11,800 training and 14,020 testing images.
Each model’s visual backbone is frozen, and a single linear head (BatchNorm + 1×1 Conv) is trained using cross-entropy loss to predict affordance maps.
We follow a consistent feature selection strategy: for each ViT-based model, four equally spaced layers are extracted and fused for probing.
Performance is measured by mean Intersection-over-Union (mIoU).
To further test the contribution of explicit 3D information, we augment each visual feature map with estimated depth and surface-normal cues from Metric3Dv2~\cite{hu2024metric3d}, spatially aligned and concatenated with the visual patches.
Six VFMs of comparable scale are evaluated under identical settings (see Table~\ref{tab:model_comparison}).

\noindent\textit{Results and Analysis.}
Fig.~\ref{fig:sec2/linear_prob} shows models with stronger geometric awareness in Probing3D~\cite{el2024probing} rankings achieve higher mIoU on UMD affordance segmentation, confirming a direct link between geometric awareness and affordance separability.
Adding Metric3Dv2~\cite{hu2024metric3d} depth and surface-normal cues consistently improves performance, especially for CLIP and SigLIP, whose representations are primarily semantic.
In contrast, DINOv2 gains little from explicit 3D cues, a finding consistent with its top ranking in Probing3D~\cite{el2024probing}, suggesting its pretraining already embeds strong geometric priors.
These results demonstrate that geometric representation forms the structural basis of affordance reasoning, motivating a closer analysis of how different VFMs internally encode geometry.

\begin{figure}[!t]
  \centering
  \includegraphics[width=\linewidth]{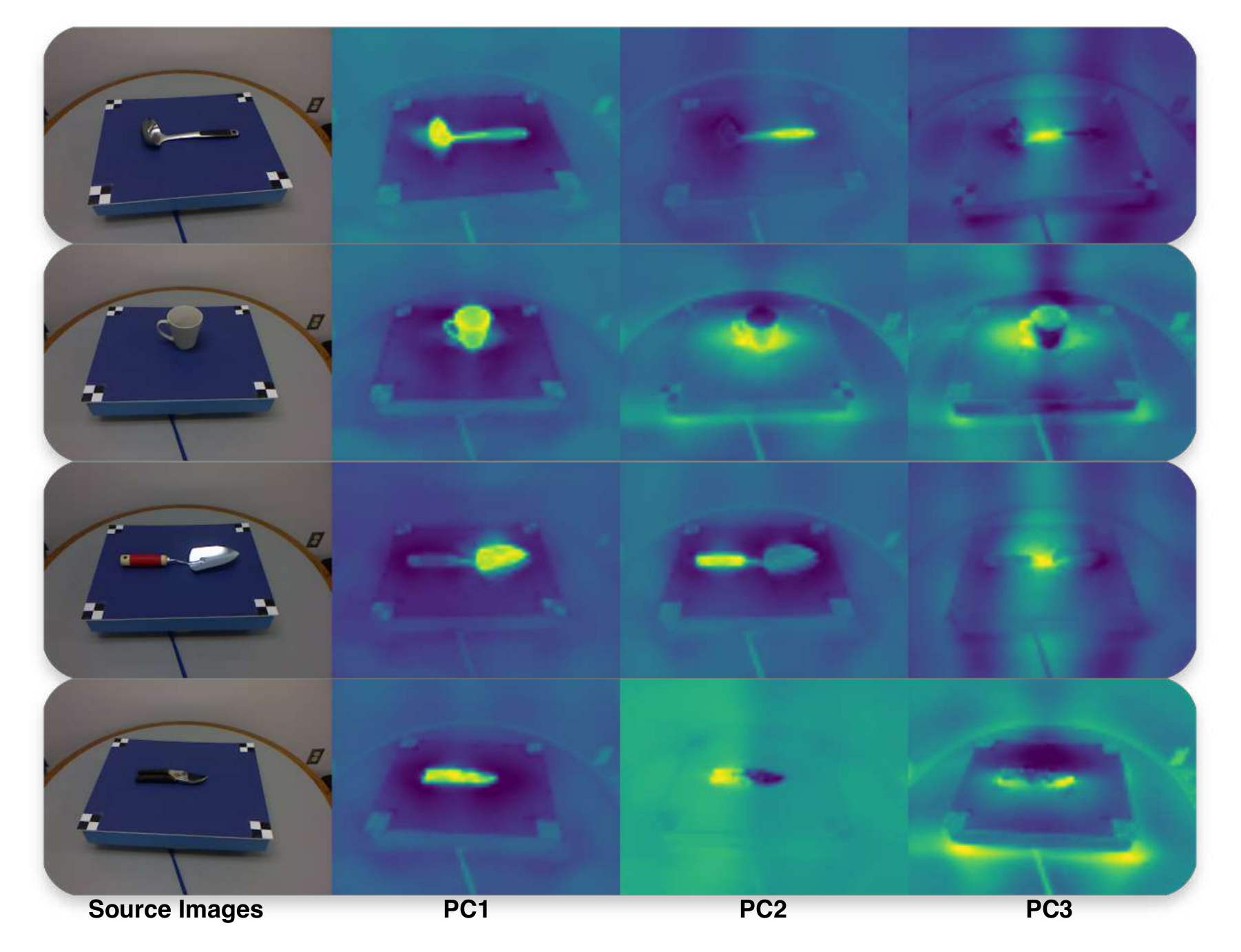}
  \caption{ \textbf{Stable part-level structure in DINO representations.} PCA of DINOv3 embeddings reveals consistent part-wise activations across the first three components, indicating interpretable geometric decomposition aligned with object functionality. }
  \label{fig:sec2/pca}
\end{figure}

\begin{figure}[!t]
  \centering
  \includegraphics[width=\linewidth]{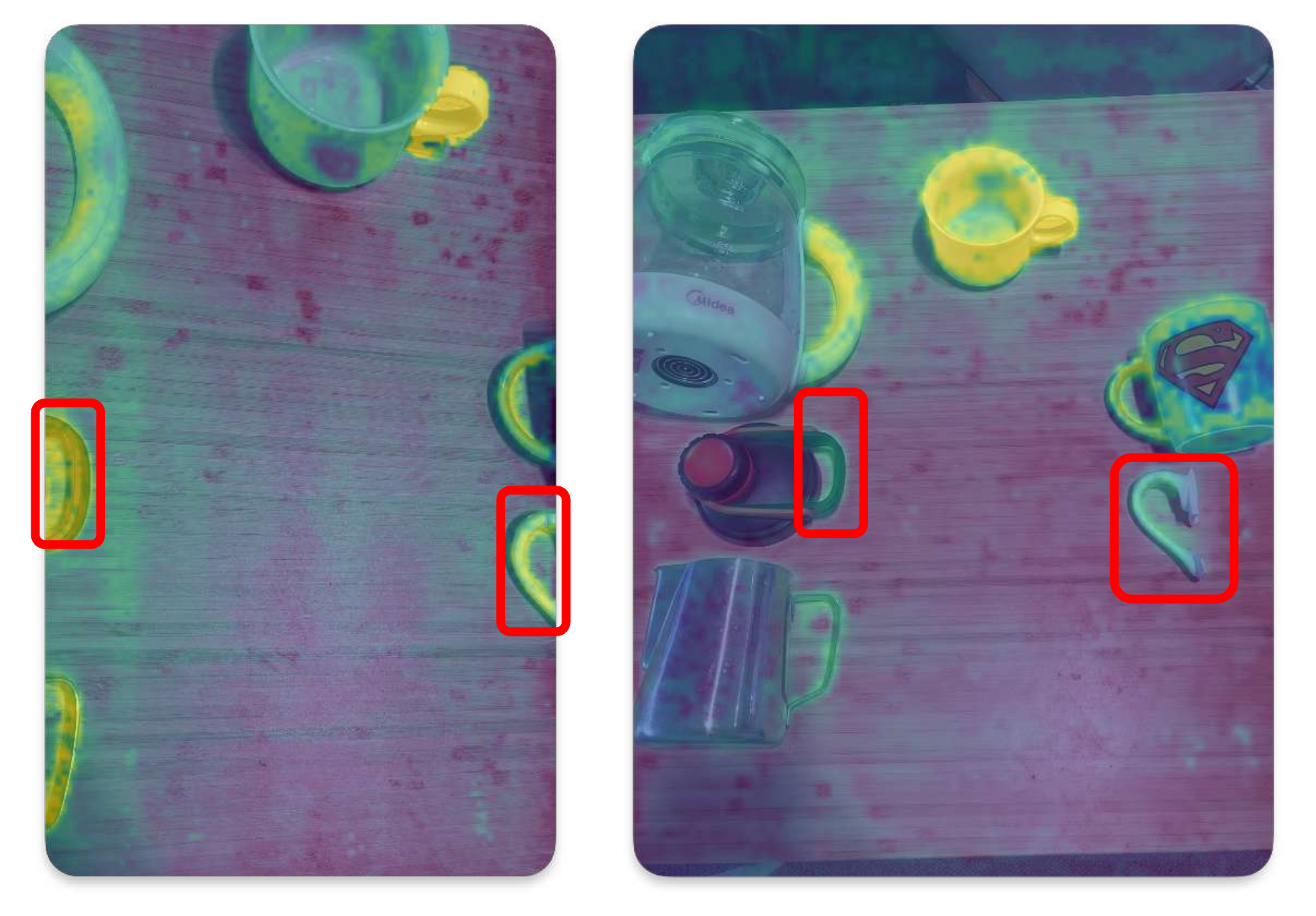}
  \caption{ \textbf{Semantic assimilation of geometry in DINO. } A mug-handle patch(Fig.~\ref{fig:sec2/cos_sim} left top) activates all ring-like shapes regardless of material or color (left), confirming that DINO’s local correspondence arises from genuine geometric similarity rather than shared semantics. When the full scene is shown (right), these responses vanish, indicating that global object semantics suppress purely geometric alignment.
  }
  \label{fig:sec2/semantic}
\end{figure}

\vspace{3pt}
\noindent\textbf{Observation 1.2. Distinct Geometric Representations Across Model Families.}
Having established a quantitative link between geometry and affordance, we proceed to qualitatively dissect how this geometry is encoded within different VFMs. By projecting diverse scenes into a unified geometric subspace, we uncover a striking divergence in representation strategies (Fig.~\ref{fig:sec2/cos_sim}). DINO families uniquely organize scenes into semantically meaningful, part-level structures—cleanly isolating handles, rims, and other functional components. This stands in stark contrast to other models: SAM reduces geometry to object boundaries, Stable Diffusion to smooth surface-like embeddings, and CLIP's mid-level geometry collapses into semantic categories in deeper layers. Among these varied strategies, DINO's explicit part-level decomposition emerges as the most compelling, given the established importance of parts in affordance theory~\cite{li2024one, xu2022partafford}.

\vspace{3pt}
\noindent\textbf{Observation 1.3. Part Decomposition and Semantic Assimilation in DINO.}
Motivated by its strong part-level organization, we analyze DINO~\cite{simeoni2025dinov3} in greater depth.
PCA on UMD object embeddings reveals that its leading components consistently correspond to meaningful parts—handles, blades, and grips—indicating a stable geometric decomposition (Fig.~\ref{fig:sec2/pca}).
However, \cite{el2024probing} argues that such part correspondence in DINO may arise from semantic association rather than geometric understanding. To disentangle this, we designed another test using a mug-handle patch as a geometric probe (Fig.~\ref{fig:sec2/cos_sim}).
We computed its cosine similarity with features extracted from two different viewpoints of the same objects: one viewpoint providing minimal semantic context about the object's identity, and another offering a complete semantic context. This paired comparison isolates how DINO's response to the same geometric structure varies with semantic information.
% ~\XS{This test is very interesting, but this explanation is confusing and it jumps from our test to the conclusion directly, (too quick), please explain more about how you designed this test, say, what difference are the two input images.}
The results reveal a nuanced picture. In simplified scenes containing only ring-like shapes, DINO's response is purely geometric—it strongly activates all ring structures, regardless of their material or color. However, in full object contexts, this geometric response is sharply attenuated for rings that are not semantically part of a cup (Fig.~\ref{fig:sec2/semantic}).

\begin{figure}[!t]
  \centering
  \includegraphics[width=\linewidth]{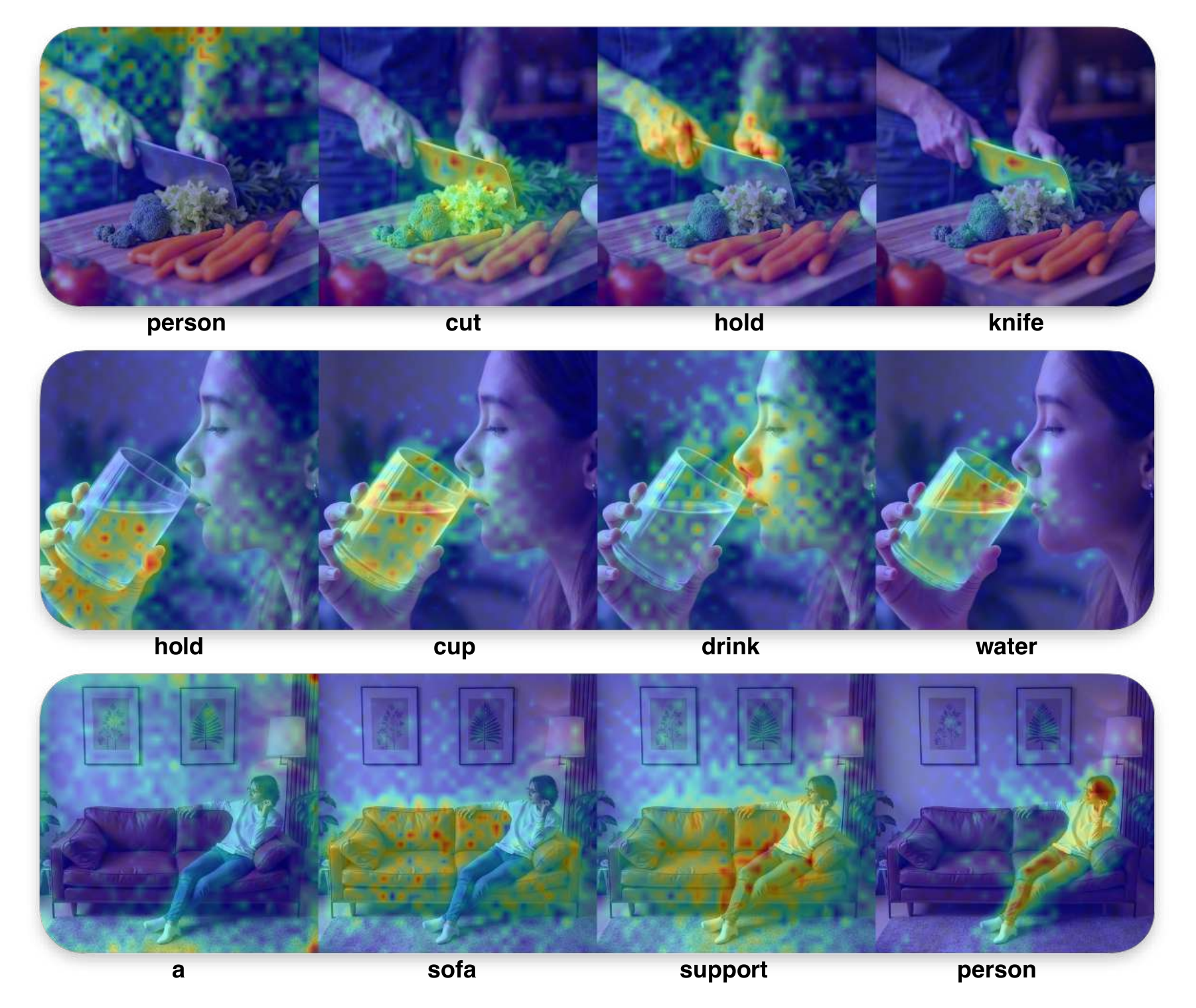}
  \caption{\textbf{Verb-conditioned attention in generative models.} Cross-attention maps show that verbs such as cut, drink, and support consistently localize to contact regions, revealing implicit interaction priors in Flux.
    }
  \label{fig:sec3/flux_cross}
\end{figure}

After establishing the correlation between geometry and affordance in VFMs, we attribute the strong affordance reasoning capability of top-performing models to their explicit object part-level representations. Our investigation reveals that DINO's representations exhibit a coupling of geometric and semantic information. To isolate more pure, shape-centric geometric knowledge at the part level, we employ PCA to extract interpretable geometric prototypes, which are subsequently utilized in our fusion pipeline.

\subsection{Interaction}
\label{sec3:interaction}

Building on our previous definition of affordance, we now analyze another dimension—interaction, describing how an agent engages to object parts.
We evaluate this ability through the verb understanding of VFMs and find that generative models, naturally encode strong interaction priors.
Their verb-conditioned cross-attention maps align verbs with corresponding functional regions and can directly serve as zero-shot predictors.

\begin{figure*}[!t]
  \centering
  \includegraphics[width=\linewidth]{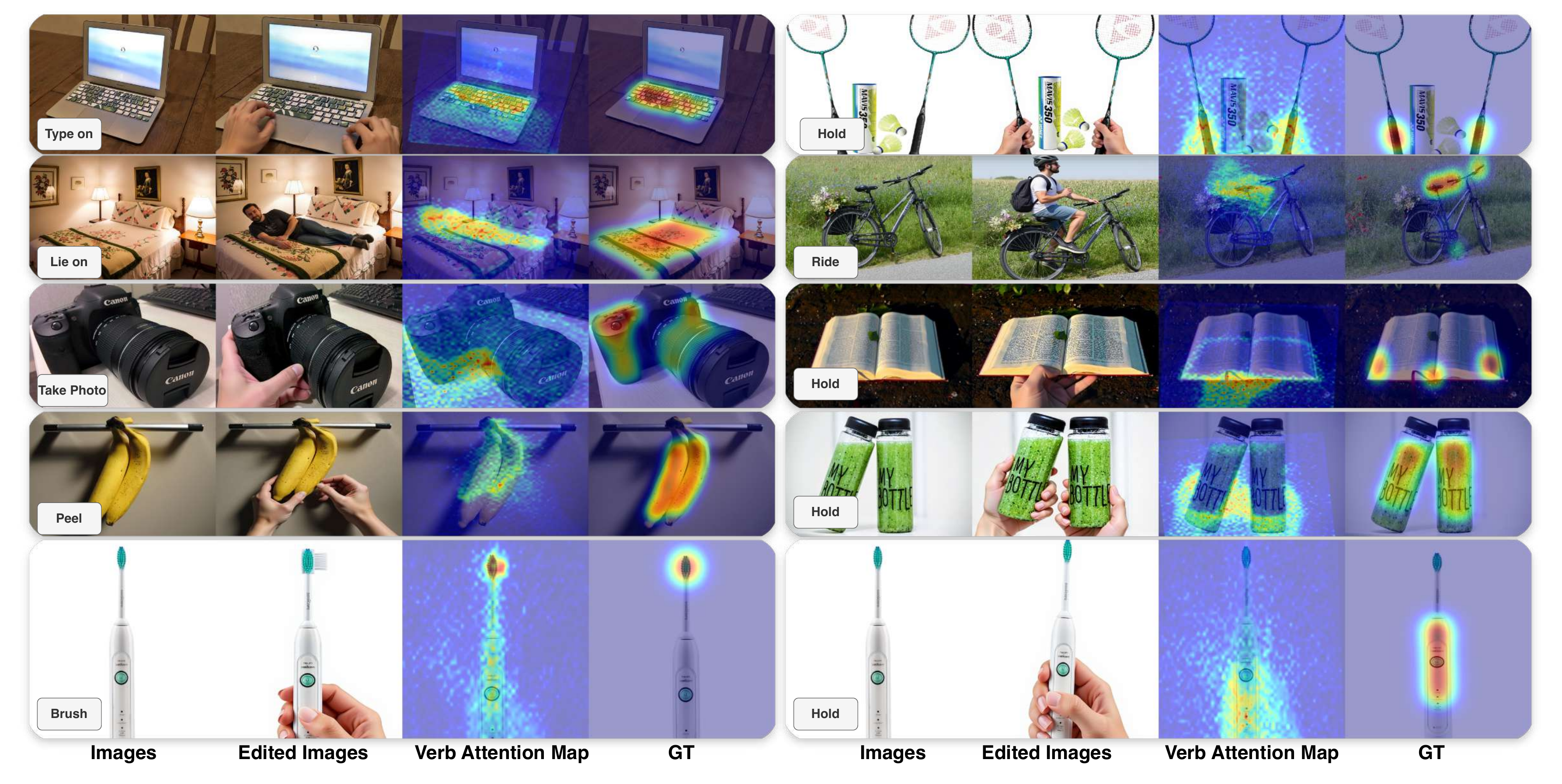}
  \caption{\textbf{Qualitative visualization of verb-conditioned interaction maps extracted by Flux Kontext.}
}
  \label{fig:sec3/main_result}
\end{figure*}

\begin{figure}[!t]
  \centering
  \includegraphics[width=\linewidth]{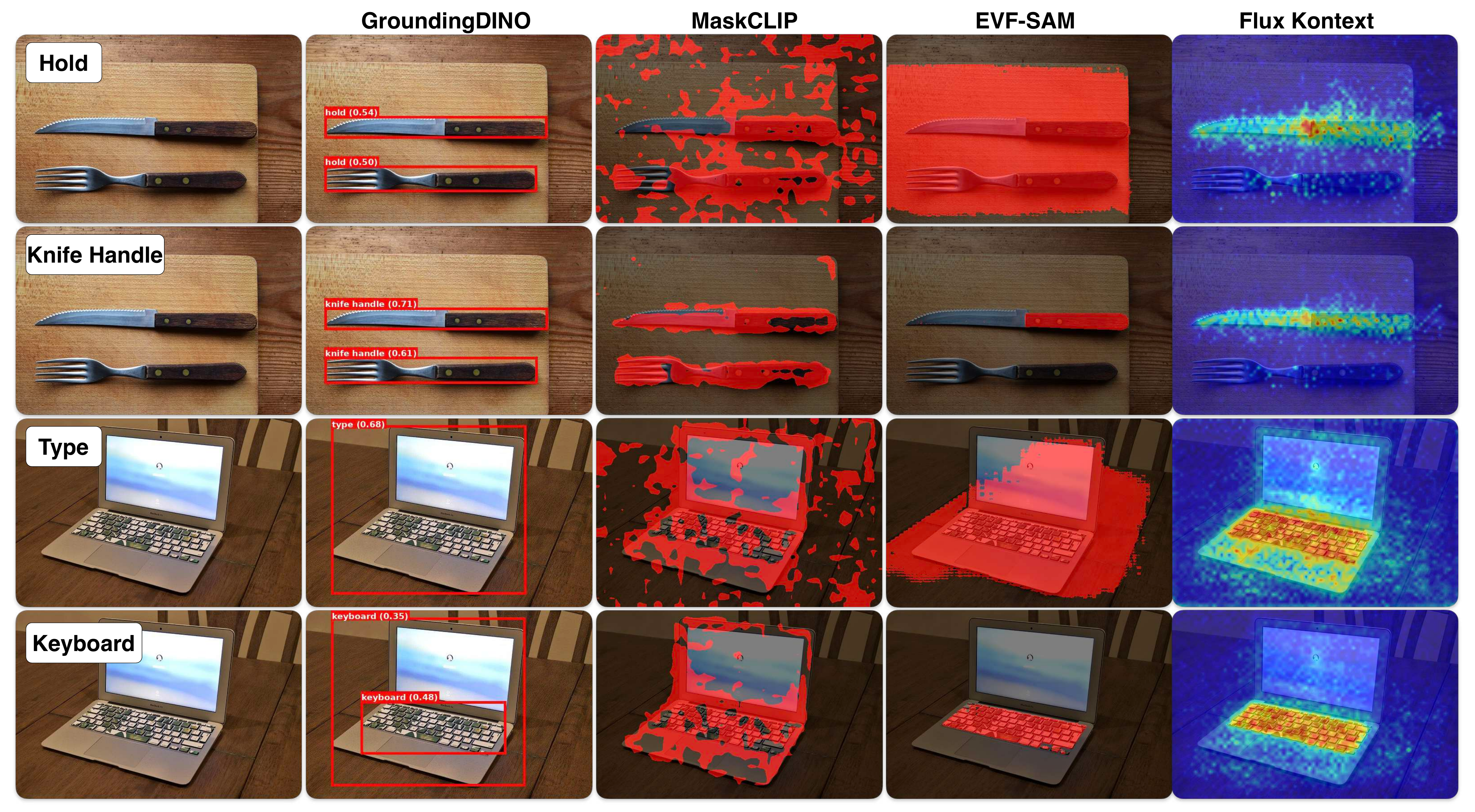}
      \caption{\textbf{Verb-conditioned localization across models.} The first and third lines are used to query verbs, and the second and fourth lines are object nouns corresponding to verbs.
    }
  \label{fig:sec3/interaction_models}
\end{figure}
\vspace{5pt}
\noindent\textbf{Observation 2.1. Emergent interaction cues from genertive VFMs.}
Weakly supervised affordance models associate actions and objects by observing human–object interactions~\cite{fang2018demo2vec,nagarajan2019grounded,liu2022joint,chen2023affordance,li2023locate,luo2022learning,qian2024affordancellm,jiang2025affordancesam}, inferring actionable regions from verb–context correlations rather than explicit geometry (Fig.~\ref{fig:sec1/paradigm}).
This raises a key question: if such priors can be learned from observation, might generative models—capable of \emph{synthesizing} interactions—already learned the interaction cues
We explore this by comparing diffusion models and find that \textbf{Flux.1-dev}~\cite{labs2025flux} exhibits the most consistent verb–scene alignment (see Appendix).
Analyzing its text–image cross-attention (Fig.~\ref{fig:sec3/flux_cross}) reveals that verbs like \textit{hold}, \textit{use}, and \textit{support} consistently attend to contact regions between agents and objects, while object tokens focus on corresponding entities.
This separable pattern suggests that generative models inherently encode \emph{verb-conditioned interaction priors}, allowing verbs to control not only \emph{what} is generated but also \emph{where}—revealing an implicit spatial organization underlying interaction perception.
\begin{figure}[!t]
  \centering
  \includegraphics[width=\linewidth]{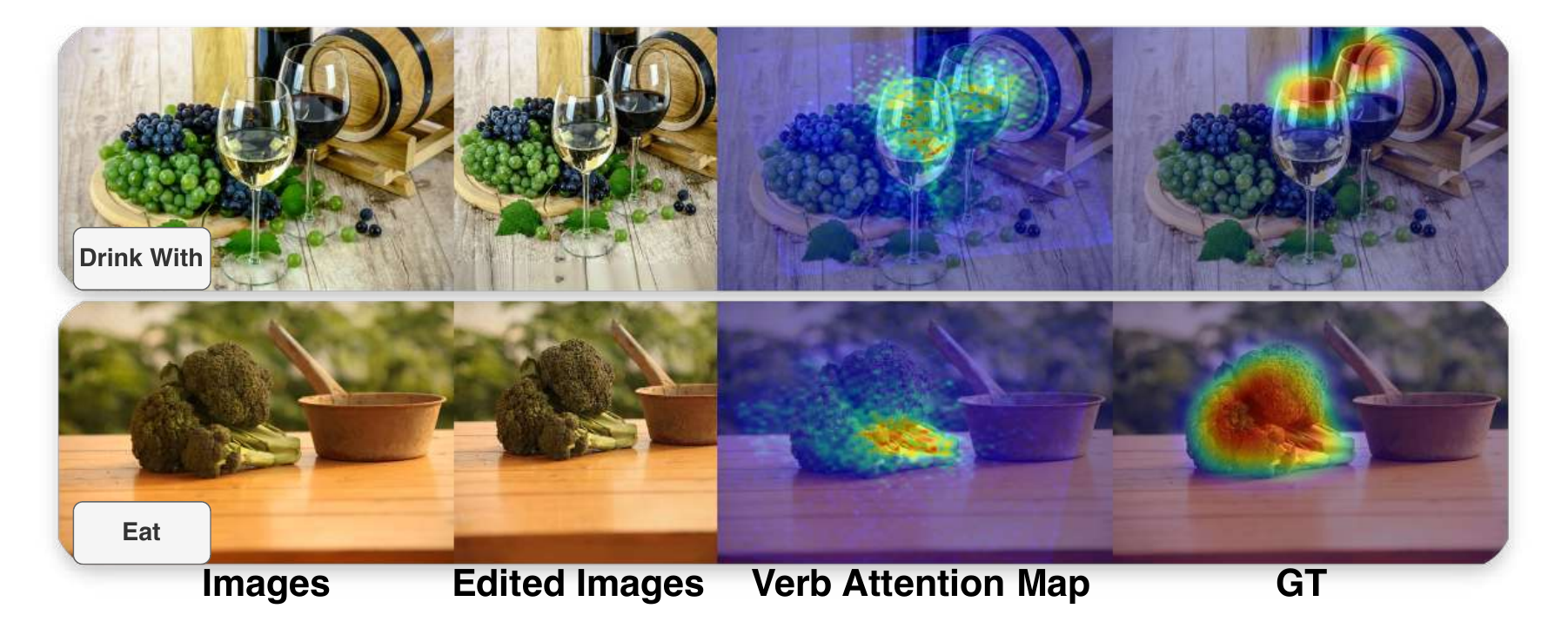}
  \caption{ \textbf{Verb attention persists despite generation failure.}
}
  \label{fig:sec3/generate_fail}
\end{figure}

 \begin{figure*}[!t]
  \centering
  \includegraphics[width=\linewidth]{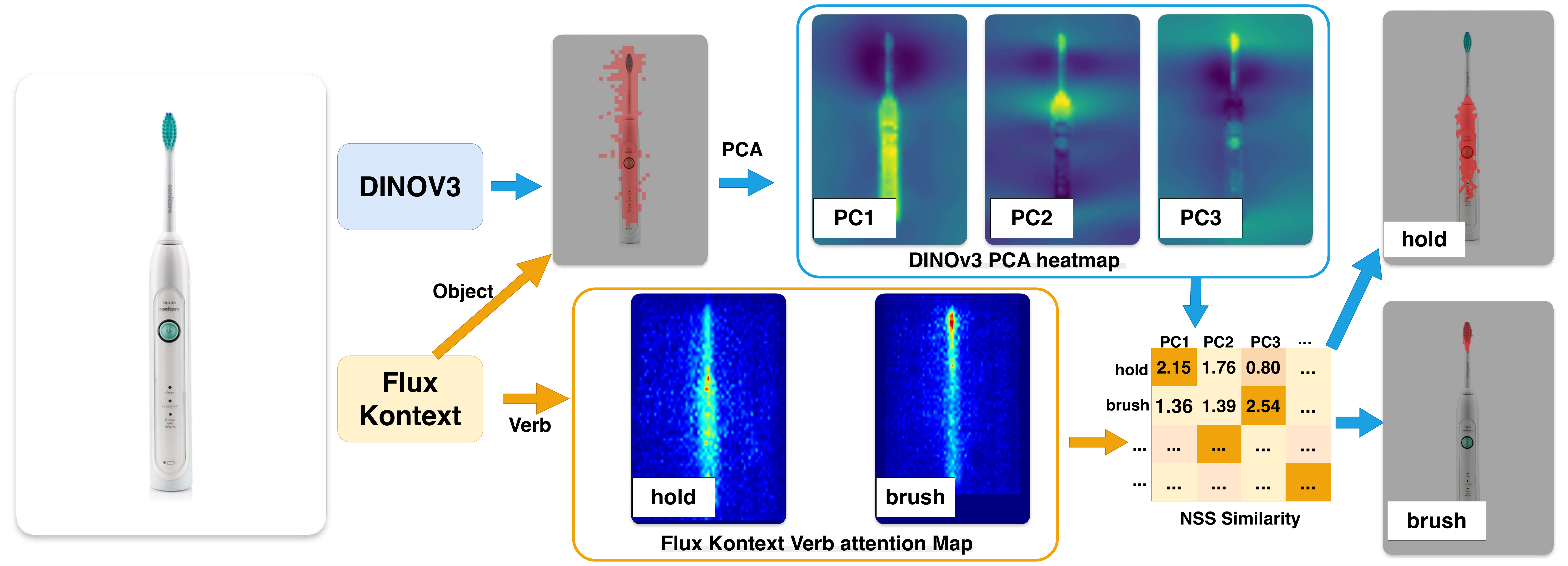}
  \caption{\textbf{Fusion pipeline of geometry and interaction.} DINOv3 captures geometric parts via PCA, Flux Kontext encodes verb-conditioned attention, and Normalized Scanpath Saliency links them to align structure with interaction semantics.
}
  \label{fig:sec4/pipeline}
\end{figure*}

% \begin{table}[t]
%   \centering
%   \setlength{\tabcolsep}{4pt}
%   \renewcommand{\arraystretch}{1.05}
%   % 使用 \resizebox 强制单栏宽度；S 列实现数值对齐且不会换行重叠
%   \resizebox{\columnwidth}{!}{%
%   \begin{tabular}{@{}l l
%       S[table-format=1.3]
%       S[table-format=1.3]
%       S[table-format=1.3]@{}}
%     \toprule
%     Method & Supervision & {KLD$\downarrow$} & {SIM$\uparrow$} & {NSS$\uparrow$} \\
%     \midrule
%     AffordanceLLM                  & Fully-supervised & 1.463 & 0.377 & 1.070 \\
%     Cross-View-AG                  & Weakly-supervised& 1.787 & 0.285 & 0.829 \\
%     LOCATE                         & Weakly-supervised & 1.405 & 0.372 & 1.157 \\
%     Affordance-R1                  & Post-Training     & 9.730 & 0.360 & 0.980 \\ \hdashline
%     Ours(Only Interaction)         & Training-Free     & 2.565 & 0.262 & 0.735 \\
%     \bottomrule
%   \end{tabular}%
%   }
%   \caption{
%     Quantitative comparison on AGD20K.
%     Our zero-shot \textbf{Flux Kontext} achieves substantially lower KLD than the prior zero-shot baseline (\textbf{Affordance-R1}),
%     approaching the weakly-supervised \textbf{LOCATE} performance.
%     This demonstrates the presence of strong implicit interaction priors in diffusion-based attention.
%   }
%   \label{tab:sec3/agd20k_results}
% \end{table}

\vspace{5pt}
\noindent\textbf{Observation 2.2. From attention to interaction understanding.}
The above observations suggest that generative models inherently encode interaction priors.
To make these implicit cues spatially explicit, we use \textbf{Flux Kontext}~\cite{labs2025flux}, a generative editing model built on Flux.
Based on this model, we design a controlled image-editing framework to extract verb-conditioned interaction cues during the generation process.
Each interaction is represented as a triplet of \textit{agent}, \textit{object}, and \textit{verb} (e.g., “add a hand to hold the knife”) (see Fig.~\ref{fig:sec3/main_result}).
Through templated prompts, the model synthesizes specific interaction scenes, from which we record verb-token cross-attention across layers.
These attention maps, obtained directly from the generative process without supervision, reveal explicit spatial distributions of interaction likelihood—making latent relational cues observable and measurable.

\vspace{5pt}
\noindent\textit{Qualitative and quantitative evaluation.}
To examine how different models encode verb–object relations, we compare representative vision–language frameworks that extend geometric encoders with semantic alignment, including \textbf{GroundingDINO}~\cite{liu2024grounding}, \textbf{MaskCLIP}~\cite{dong2023maskclip}, and \textbf{EVF-SAM}~\cite{zhang2024evf} (Fig.~\ref{fig:sec3/interaction_models}).
EVF-SAM shows clear part-level segmentation but diffuse verb attention, while GroundingDINO and MaskCLIP capture global semantics yet fail to localize functional regions.
In contrast, \textbf{Flux Kontext} achieves coherent alignment between verbs and action-bearing parts, motivating our focus on its interaction priors in the following analyses.

Verb-conditioned attention maps extracted via \textbf{Flux Kontext} (Fig.~\ref{fig:sec3/main_result}) consistently highlight physically plausible contact regions—such as hands when \textit{typing} on a laptop or grasped areas when \textit{holding} a bottle.
Even in failed generation cases where agents (Fig.~\ref{fig:sec3/generate_fail}), the verb attention remains spatially consistent, indicating that interaction priors originate from internal  rather than pixel-level tracking.
For quantitative evaluation, we treat these attention maps as zero-shot affordance predictions on \textbf{AGD20K}~\cite{luo2022learning}, 
As summarized in Table~\ref{tab:sec3/agd20k_results},  it's approaching weakly supervised performance despite requiring no affordance-specific training.
These results confirm that diffusion-based attention inherently captures verb-conditioned spatial organization aligned with human affordance perception, bridging the geometric and interactional foundations of actionable understanding.

\begin{table}[t]
  \centering
  \setlength{\tabcolsep}{4pt}
  \renewcommand{\arraystretch}{1.05}
  % 使用 \resizebox 强制单栏宽度；S 列实现数值对齐且不会换行重叠
  \resizebox{\columnwidth}{!}{%
  \begin{tabular}{@{}l l
      S[table-format=1.3]
      S[table-format=1.3]
      S[table-format=1.3]@{}}
    \toprule
    Method & Supervision & {KLD$\downarrow$} & {SIM$\uparrow$} & {NSS$\uparrow$} \\
    \midrule
    AffordanceLLM                  & Fully-supervised & 1.463 & 0.377 & 1.070 \\
    Cross-View-AG                  & Weakly-supervised& 1.787 & 0.285 & 0.829 \\
    LOCATE                         & Weakly-supervised & 1.405 & 0.372 & 1.157 \\
    Affordance-R1                  & Post-Training     & 9.730 & 0.360 & 0.980 \\ \hdashline
    Ours(Only Interaction)         & Training-Free     & 1.825 & 0.271 & 1.050 \\
    Ours(Interaction $\times$ Geometry)   & Training-Free     & 1.493 & 0.326 & 1.090 \\
    \bottomrule
  \end{tabular}%
  }
  \caption{ \textbf{Quantitative comparison on AGD20K~\cite{luo2022learning}.}
    We evaluate our zero-shot, training-free approach against both fully- and weakly-supervised baselines. Using only verb attention from Flux Kontext achieves competitive scores.
  }
  \label{tab:sec3/agd20k_results}
\end{table}
\section{Geometry-Interaction Integrated Affordance Estimation}
\label{sec:fusion}
Building on our findings of part-level geometry in discriminative models (Sec.~\ref{sec3: Geometry})  and verb-conditioned priors in generative models (Sec.~\ref{sec3:interaction}), we now present a definitive test of our framework: a unified and training-free fusion approach. We posit that if these two dimensions are fundamental to affordance, their explicit composition should enable affordance estimation without any supervision. We first detail the fusion procedure, then validate this hypothesis quantitatively and qualitatively, concluding with a discussion of limitations.

% ------------------------------
\subsection{Fusion Procedure}
\label{sec:fusion_method}

To demonstrate that geometry and interaction can be explicitly composed into functional affordance reasoning,
we construct a simple training-free pipeline using \textbf{DINOv3}~\cite{simeoni2025dinov3} and \textbf{Flux Kontext} (Fig.~\ref{fig:sec4/pipeline}).
We adopt DINOv3 for its stronger spatial coherence and compact part-level features.
For a given image, DINOv3 provides dense representations capturing object geometry,
while Flux Kontext generates verb and object conditioned cross-attention maps encoding interaction cues.
The object attention is used to crop a region of interest (ROI) from the DINOv3 feature map,
and the ROI features are decomposed into part-level bases via PCA.
In parallel, the verb attention map indicates the probable action region.
We compute Normalized Scanpath Saliency(NSS) between each geometric basis and the verb attention,
select the best-aligned component, and fuse it with the verb attention to produce a verb-specific affordance mask.
This process directly operationalizes the complementary dimensions—geometry and interaction—into actionable, interpretable spatial predictions,
achieved entirely without fine-tuning or task-specific supervision.
% ----------

\subsection{Quantitative and Qualitative Evaluation}
\label{sec:fusion_results}

\noindent\textbf{Quantitative results.}  
We evaluate our framework on the AGD20K~\cite{luo2022learning}, a standard dataset for weakly-supervised affordance estimation. It contains egocentric-view images of objects paired with exocentric-view interaction exemplars.  We evaluate directly on its test set of unseen ego-centric objects, without leveraging any of the dataset's training interaction data. As shown in Table~\ref{tab:sec3/agd20k_results}, our full Interaction × Geometry method consistently outperforms the Only Interaction baseline across metrics. The significant reduction in KLD indicates that geometric cues yield sharper, more confident predictions by suppressing implausible regions. Concurrent gains in SIM and NSS confirm that the fused masks better cover the ground-truth distribution while highlighting the most salient action areas. We quantitatively evaluate on AGD20K because its interaction heatmaps align with our verb-conditioned outputs. UMD provides categorical masks, so we use it only for qualitative validation of geometric coherence.
% \JZ{we don't have any ablation study? e.g. PCA?}

\begin{figure}[!t]
  \centering
  \includegraphics[width=\linewidth]{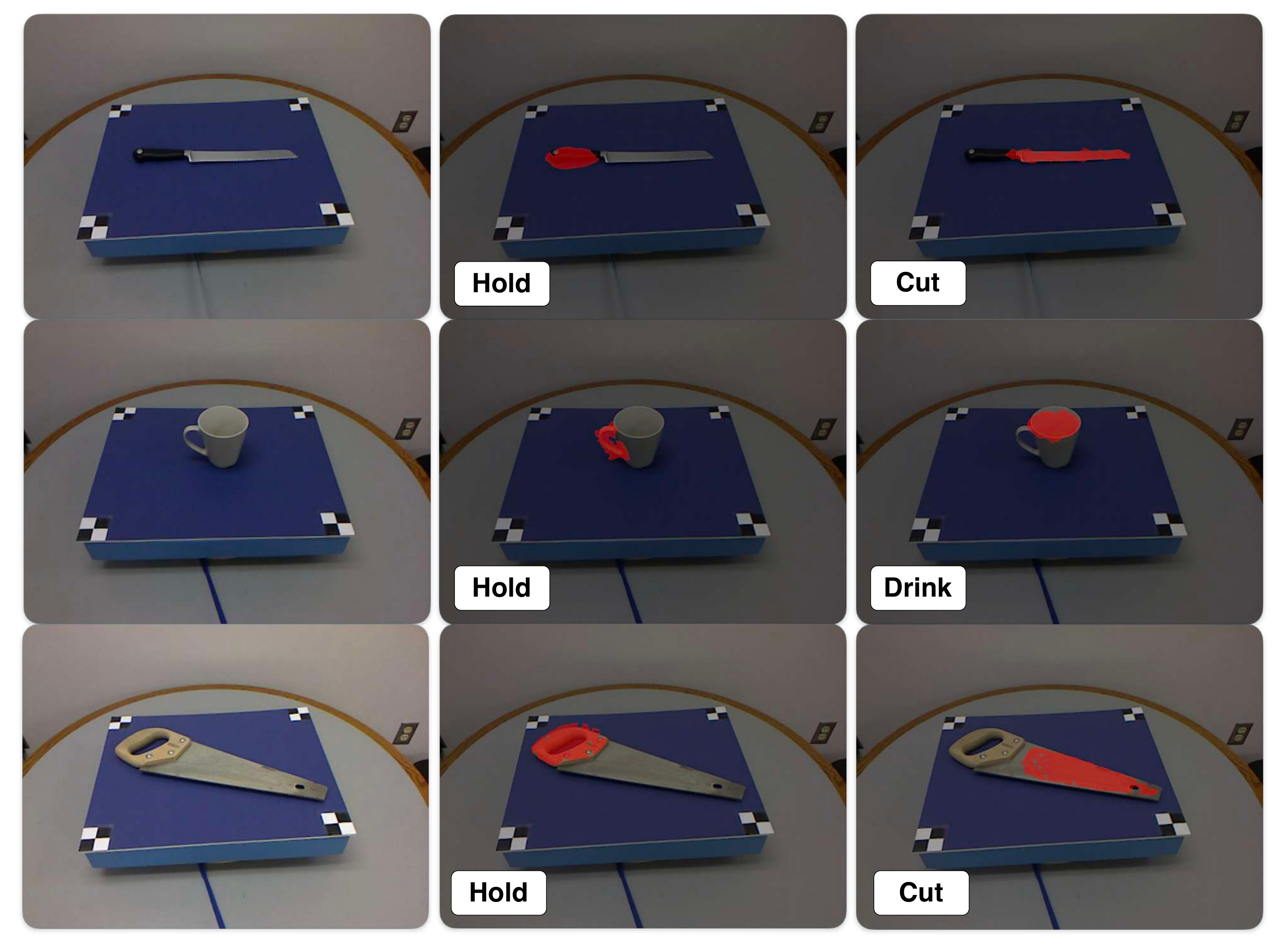}
  \caption{ \textbf{Affordance segmentation from geometry–interaction fusion.} The fused model localizes plausible action regions (e.g., handles, blades, rims) on the UMD dataset—without affordance-specific training.}
  
  % \XS{The qualitative results about the comparison among geometry–interaction, geometry-only, and interaction-only, will be more convincing, if time allows. }
    
  \label{fig:sec4/fusion_vis}
\end{figure}

\noindent\textbf{Qualitative results.}  
% To visualize the behavior of the proposed fusion,
We apply the proposed fusion strategy to representative examples from the UMD dataset~\cite{do2018affordancenet} (Fig.~\ref{fig:sec4/fusion_vis}) to further explain its effectiveness.  
Across different object categories and actions, the composed affordance masks align well with functional parts:  
\textit{hold} regions localize to handles,  
\textit{cut} regions to blades, and  
\textit{drink} regions to cup openings.  
These qualitative examples provide intuitive evidence that the fusion mechanism effectively binds geometric structure and action semantics into coherent, interpretable affordance regions.
% ~\XS {Given that we have conducted the quantitative experiment on UMD when analysing geometric prior. The reviewer might wonder why we don't do the quantitative analyssi on UMD dataset as well, so it might provide a short justification why we only do the qualitative analysis on UMD.}

\subsection{Discussion}
\label{sec:fusion_discussion}

\noindent\textbf{Benefits of the two worlds:} 
Our framework decomposes affordance into geometric and interaction capacities, which can reside in separate models. While instantiated with DINOv3 and Flux, the paradigm is general: any model with strong part-level geometry can be paired with any model providing verb-conditioned spatial priors for zero-shot reasoning. This shifts the focus from training specific models to composing existing capabilities.

% \JZ{discuss our choice is not the only solution. Explain that our observation provide a new direction in performing affordance estimation, e.g. any model with higher degree of modeling geometry + another model capable of modeling interaction can be a good choice.}

\noindent\textbf{IS PCA necessary?} 
We use PCA to address the geometric-semantic entanglement in DINO's features as discovered in Fig.~\ref{fig:sec2/semantic}, where using features directly would introduce excessive noise. Essentially, our primary objective is to extract compact, part-level geometric prototypes, which also points toward the future challenge of obtaining purer, shape-centric geometric knowledge from Visual Models.

% \JZ{explain why you perform PCA and if there are other alternatives}

\noindent\textbf{Video models can bring extra benefits:} 
Looking forward, video generative models offer a promising direction by unifying our two dimensions. To generate physically plausible interactions, they must develop an inherent understanding of 3D scene geometry alongside interaction dynamics. This makes them ideal unified sources for both spatial geometric priors and interaction priors, providing a more complete foundation for affordance reasoning in dynamic environments.

% \JZ{discuss potential of using video generative models, e.g. video world models for affordance, as they can contain both geometry information and interaction }

\noindent\textbf{Limitation analysis:}
The efficacy of our current paradigm is constrained by the quality of the primitives we extract—such as the noisiness of noun-conditioned attention maps and the instability of generative outputs—and by the simplicity of our training-free fusion strategy, which performs a shallow, open-loop signal combination.

% The most promising direction involves layered reasoning architectures for open-vocabulary natural language instructions. The key challenge lies in distinguishing and synergizing functionality - an object's intended purpose (e.g., \enquote{turning lights on/off}) where LLMs excel - from affordance - physical interaction possibilities (e.g., \enquote{pressing/rotating}) addressed by our geometric-interaction framework.

\section{Conclusion}

% --------------------- 5.1 ---------------------

% --------------------- 5.2 ---------------------
% \subsection{Conclusion}

This work demonstrates that a model truly understands affordance when it represents two complementary dimensions: the geometry of object structures that support action, and the interaction dynamics between an agent and those structures. We introduced a dual-dimensional framework to formalize this view, systematically probed how these capabilities are embedded within diverse Visual Foundation Models, and validated their complementarity through a training-free fusion approach that achieves zero-shot affordance estimation.
Our findings reveal that part-level geometric organization and verb-conditioned interaction priors exist as composable elements in pre-trained models. By showing that these primitives can be explicitly mined and fused without supervision, we establish a new paradigm for affordance reasoning—one that moves from training task-specific models to assembling innate capabilities. This path forward, focused on refining and composing pre-trained primitives, offers a promising foundation for more general and actionable visual understanding in embodied intelligence systems.

% The most promising direction involves layered reasoning architectures for open-vocabulary natural language instructions. The key challenge lies in distinguishing and synergizing functionality - an object's intended purpose (e.g., \enquote{turning lights on/off}) where LLMs excel - from affordance - physical interaction possibilities (e.g., \enquote{pressing/rotating}) addressed by our geometric-interaction framework.


{
    \small
    \bibliographystyle{ieeenat_fullname}
    \bibliography{main}
}
%\input{sec/X_review_pipeline}


\end{document}